\documentclass[conference]{IEEEtran}
\IEEEoverridecommandlockouts

\usepackage{cite}
\usepackage{amsmath,amssymb,amsfonts}
\usepackage{algorithm}
\usepackage{algpseudocode}
\usepackage{graphicx}
\usepackage{textcomp}
\usepackage{xcolor}
\usepackage{braket}
\usepackage{csquotes} 
\usepackage{mathtools}
\usepackage{multirow}
\usepackage{relsize}
\usepackage{caption}
\usepackage{subcaption}
\usepackage{xcolor}
\usepackage{hyperref}
\usepackage{tikz}
\usepackage{blkarray}
\usepackage{hf-tikz}
\usepackage{physics}
\usepackage{colortbl}
\usepackage{multicol}

\usepackage{booktabs}
\usepackage{nicefrac}
\usepackage{complexity}
\usepackage{listings}
\usepackage{graphicx}
\usepackage{lscape}
\usepackage{amsthm}
\usepackage{mathrsfs} 
\usetikzlibrary{decorations.pathreplacing, calc, automata, matrix, quantikz2, positioning, trees, decorations.pathmorphing, shapes}
\def\BibTeX{{\rm B\kern-.05em{\sc i\kern-.025em b}\kern-.08em
    T\kern-.1667em\lower.7ex\hbox{E}\kern-.125emX}}

\newtheorem{theorem}{Theorem}[]
\newtheorem{lemma}{Lemma}[section]

\theoremstyle{definition}
\newtheorem{definition}{Definition}[section]

\begin{document}

\title{CNOT Minimal Circuit Synthesis: \\A Reinforcement Learning Approach
\thanks{This work is partially supported by PNRR project SERICS---Security and Rights in the CyberSpace (CUP H73C22000890001).}}



\author{\IEEEauthorblockN{Riccardo Romanello\IEEEauthorrefmark{1}, Daniele Lizzio Bosco\IEEEauthorrefmark{2}\IEEEauthorrefmark{3}, Jacopo Cossio\IEEEauthorrefmark{2}, Dusan Sutulovic\IEEEauthorrefmark{2}, \\Giuseppe Serra\IEEEauthorrefmark{2}, Carla Piazza\IEEEauthorrefmark{2} and Paolo Burelli\IEEEauthorrefmark{4}}

\IEEEauthorblockA{
\IEEEauthorrefmark{1}\textit{Department of Environmental Sciences, Informatics and Statistics}\\ 
\textit{Ca' Foscari University of Venice, Venice, Italy}\\
\IEEEauthorblockA{
\IEEEauthorrefmark{2}\textit{Department of Mathematics, Computer Science and Physics}\\
\textit{University of Udine, Udine, Italy}}
\IEEEauthorblockA{
\IEEEauthorrefmark{3}\textit{Department of Biology}\\
\textit{University of Naples Federico II, Naples, Italy}}
\IEEEauthorblockA{
\IEEEauthorrefmark{4}\textit{Play, Culture and AI Section}\\
\textit{IT University of Copenhagen, Copenhagen, Denmark}}
}
Email: riccardo.romanello@unive.it, lizziobosco.daniele@spes.uniud.it
}

\maketitle

\begin{abstract}
CNOT gates are fundamental to quantum computing, as they facilitate entanglement, a crucial resource for quantum algorithms. Certain classes of quantum circuits are constructed exclusively from CNOT gates. Given their widespread use, it is imperative to minimise the number of CNOT gates employed.
This problem, known as \textit{CNOT minimization}, remains an open challenge, with its computational complexity yet to be fully characterized.
In this work, we introduce a novel reinforcement learning approach to address this task.
Instead of training multiple reinforcement learning agents for different circuit sizes, we use a single agent up to a fixed size $m$.
Matrices of sizes different from m are preprocessed using either \textit{embedding} or \textit{Gaussian striping}. To assess the efficacy of our approach we trained an agent with $m = 8$, and evaluated it on matrices of size n that ranges from $3$ to $15$. The results we obtained show that our method overperforms the state of the art algorithm as the value of n increases.
\end{abstract}


\section{Introduction}

Quantum computing is a field at the intersection of physics and computer science. It explores the properties, limitations, and capabilities of computing devices whose internal operations are governed by the rules of quantum mechanics---quantum computers.
With the term quantum computing paradigm, we refer to the set of principles and operations that define how quantum algorithms are constructed and executed.
The most widely adopted quantum computing paradigm is the gate-based model, where quantum operations are represented by unitary matrices that, in turn, have a one to one correspondence to the so-called \emph{gates}. However, practical quantum hardware imposes constraints on the direct implementation of many gates, necessitating their decomposition into smaller, hardware-compliant operations. This process is known as \emph{quantum circuit synthesis} 
\cite{kliuchnikov2013synthesisunitariescliffordtcircuits,Amy_2013,paradis2024synthetiq}
Quantum circuit synthesis is computationally demanding, as an $n$-qubit gate corresponds to a $2^n \times 2^n$ complex unitary matrix. Despite its general complexity, various subproblems within synthesis have been actively investigated, particularly those focusing on minimizing the occurrence of specific gates in a given universal gate set. A notable case is the minimization of the $T$-gate \cite{Mosca_2021,Amy_2019,gosset2013algorithmtcount}, a critical component for quantum advantage. The classical simulation of a quantum circuit is known to be exponential in the number of $T$-gates, making its reduction a fundamental objective \cite{dawson2005solovaykitaevalgorithm,NielsenChuang}.

Another crucial operation in quantum computing is the Controlled-NOT (CNOT) gate. It is the only two-qubits gate in the Clifford+T set --- one of the most popular choices for universal gate set --- and is essential for achieving quantum entanglement. Reducing the number of CNOT gates is crucial for two reasons: (1) CNOT operations introduce significant noise due to their two-qubit interactions, and (2) they are a fundamental component of stabilizer circuits \cite{gottesman1997stabilizer}, which play a key role in quantum error correction \cite{Mondal_2024}. Notably, stabilizer circuits exhibit a canonical form \cite{aaronson2004improved}, where some computational blocks consist solely of CNOT operations. 
Hence, the length of a stabilizer circuit is deeply related to the length of these CNOT-only blocks, proving that minimising their length is a critical and relevant problem.

This leads to the \emph{CNOT minimization problem}, which seeks to find an equivalent circuit with the minimal number of CNOT operations. This problem has been addressed in three different flavours: (i)  any qubit can interact with any qubit; (ii) qubits are first mapped onto a \emph{topology graph} $G$, and CNOTs are selected only among pair of qubits adjacent in $G$; (iii) ancilla qubits can be used during computation.  

The \textit{unconstrained} case can be formulated as follows.


Each CNOT gate can be represented as a binary matrix that is equal to the identity matrix except for a single off-diagonal entry indicating the qubits affected by the operation. Given an invertible binary matrix $M$ and an elementary matrix $E_{i,j}$ corresponding to a CNOT between qubits $i$ and $j$, applying the CNOT operation corresponds to the matrix multiplication $M \leftarrow E_{i,j} M$. This transformation replaces row $i$ of $M$ with the bitwise XOR of rows $i$ and $j$. Consequently, solving the CNOT minimization problem for a given matrix $M$ reduces to finding the shortest sequence of such transformations that transforms $M$ into the identity matrix. From this perspective, the task can be identified as a \textit{planning problem}. Moreover, unlike general quantum synthesis, which operates on $2^n \times 2^n$ complex matrices, CNOT synthesis exclusively deals with matrices that have linear size with respect to $n$. This arises from the one-to-one correspondence between $n$-qubit CNOT circuits and $n \times n$ binary invertible matrices. 

In this paper, we leverage these properties to address the unconstrained CNOT minimization problem with a Reinforcement Learning (RL) approach.

RL is a branch of Machine Learning that has shown remarkable effectiveness across a wide range of application domains, including  autonomous control, strategic decision-making in board games, and even fine-tuning tasks in natural language processing.  These techniques are based on the idea of \textit{agents} that can act, by effectuating a choice between a set of possible \textit{actions}, on a given \textit{environment}, representing the current state of a problem. These actions are then (either positively or negatively) rewarded, based on their effect on the environment. Therefore, each RL technique can be described as the training of an agent that aims to maximize a reward.

The formulation of the CNOT minimization problems aligns naturally with RL, as: (i) the initial state of the environment is the starting matrix $M$; (ii) the set of actions corresponds to the possible CNOT operations; (iii) the maximal reward corresponds to finding a correct plan (as a sequence of actions) that leads to the identity matrix.  

A general limitation of RL-based solutions is that every modification in the setting requires training a new agent, which is usually computationally expensive. Moreover, as the number of possible moves increases, so does the number of training samples required, highlighting scaling limitations.

To address this issue, in this work we combine an RL approach with matrix manipulations ---specifically, sub-matrix embeddings and a greedy algorithm to ``simplify" certain rows and columns --- to enable the selection of a model trained on a given dimension $m$. When the model is applied to a matrix of a dimension lower or higher than $m$, the matrix is manipulated, allowing the initial model to find a suitable solution to the problem.
We demonstrate the efficacy of our model, trained on $m=8$, in addressing tasks of sizes ranging from $3$ to $15$, showing a clear advantage over the state-of-the-art heuristic algorithm Patel-Markov-Hayes \cite{PatelMarkovHayes}  in most cases.

To ease further research in this area, we provide the code necessary to reproduce our experiments, including the agent and the algorithms required for matrices manipulation \footnote{\href{https://github.com/RiccardoRomanello/CNOTRL}{https://github.com/RiccardoRomanello/CNOTRL}}.

The paper is structured as follows.
Firstly, in Section \ref{sec::related}, we introduce the reader to the state of the art for what concerns CNOT minimal circuits synthesis. 
The goal of Section \ref{sec::back} is to describe all the preliminaries required to fully grasp the remainder of the paper. After that, we use Section \ref{sec::methods} to provide all the details about the proposed approach to address the CNOT minimization problem. 
The tests conducted to assess the efficiency of our method and the results we obtained are described in Section \ref{sec::exp_eval}.
We draw some conclusions and propose potential follow-ups in Section \ref{sec::concl}.

\section{Related Works}
\label{sec::related}

Quantum circuit synthesis has become an increasingly significant problem since the advent of physically operational quantum computers. A general synthesis algorithm takes as input a unitary matrix $U \in \mathbb{C}^{2^n \times 2^n}$ and a universal gate set $\mathcal{B}$---with Clifford+T being the standard choice. The algorithm must then output a circuit composed solely of gates from $\mathcal{B}$ that implements $U$ exactly or approximately. 

Naturally, not all circuits are of equal interest. The primary objectives in quantum synthesis are: (i) minimizing circuit \emph{depth} to mitigate decoherence effects due to the fragile quantum state, and (ii) minimizing circuit \emph{width} to limit the number of required qubits. 

Comprehensive studies on quantum circuit synthesis in its broadest scope can be found in \cite{Giles_2013,kliuchnikov2013,Kliuchnikov_2013,Mosca_2021}. 
A Reinforcement Learning approach has been adopted for the synthesis of unitary matrices in its broadest scope in \cite{weiden2025highprecisionmultiqubitcliffordtsynthesis,kremer2025practicalefficientquantumcircuit}.

However, synthesis of Clifford+T circuits can be tackle in terms of subproblems, including the minimization of specific gate types within the universal set. One such problem is the \emph{CNOT gate minimization}. 

An algebraic approach to CNOT minimization is presented in \cite{CircsOfCNOT}, where the authors first describe the group structure underlying quantum circuits composed of CNOT gates. These structural insights are then leveraged to devise heuristics aimed at reducing the number of CNOT operations in a given circuit.

A different perspective based on \emph{Steiner Trees} is explored in \cite{CNOTNisq}. This approach considers the practical constraints imposed by hardware connectivity, as real-world quantum devices do not typically support all-to-all qubit interactions. In this context, a CNOT gate between two qubits is feasible only if the qubits are physically connected. If they are not, additional \emph{swap} gates—equivalent to three CNOT operations—must be introduced, further increasing circuit depth.

Another line of research tackles the problem using tools coming from logic programming. 
On the one hand, we have works like \cite{Meuli2018SATbasedT,shaik2024optimallayoutawarecnotcircuit} that encode the problem as SAT instance. 
On the other hand, in \cite{piazza2023asp, piazza2023synthesis} authors propose a solution to both the unconstrained and constrained version with an Answer Set Programming approach.

Regarding the application of Reinforcement Learning (RL) to quantum circuit synthesis, we highlight three key studies. Two of them \cite{weiden2024highprecisionfaulttolerantquantum,vanderlinde2023qgymgymtrainingbenchmarking} focus on training RL agents to solve the general quantum circuit synthesis problem. Despite their promising results, the task does not allow for an efficient representation, and is subjected to strict scaling limitations.  

The third study, \cite{Aaronson2019ConstrainedQC}, specifically addresses CNOT minimization using RL. However, a crucial distinction between this work and our approach is that \cite{Aaronson2019ConstrainedQC} considers the \emph{constrained} variant of the problem, obtaining a different set of actions. Moreover, different topologies were addressed by different agents. Finally, each setting had at most $4$ qubits, while in our proposed method we use up to $15$ qubits.

Our work builds upon these studies by leveraging Reinforcement Learning for unconstrained CNOT minimization while integrating Answer Set Programming to preprocess larger instances, thus enabling the RL agent to operate effectively on manageable problem sizes.

\section{Background}
\label{sec::back}

\subsection{Setting the Context: CNOT-Minimization Problem}

Modern gate-based Quantum Computers have two important limitations: 
\begin{itemize}
    \item Only a finite set of operations can be applied to the qubit 
    \item The quantum coherence can be maintained for a short amount of time
\end{itemize}
The finite set of operations a Quantum Computer can apply is usually called \emph{Universal Set of Gates} $\mathcal{B}$.
Quantum Algorithms are usually described as a single $2^n \times 2^n$ unitary matrix $\mathcal{U}$.

Rewriting $\mathcal{U}$ in terms of gates in $\mathcal{B}$ is called \emph{synthesis}.

The most common choice for $\mathcal{B}$ is the Clifford + T set, which contains the single-qubit $H$, $S$, and $T$ gates,
and the two-qubit gate CNOT: 
\begin{equation}
    \label{eq::graphs-reducts_circ-synth::CNOT}
     CNOT = \begin{pmatrix}
         1 & 0 & 0 & 0 \\
         0 & 1 & 0 & 0 \\
         0 & 0 & 0 & 1 \\
         0 & 0 & 1 & 0 \\
     \end{pmatrix}
\end{equation}

Single qubit gates can be implemented in physical quantum computers with no errors. On the other hand, two-qubit gates introduce errors that must be corrected using Quantum Error Correction (QEC) techniques. 
Hence, it is reasonable to think about synthesis techniques that aim at minimizing the number of CNOT gates in the resulting circuit. 

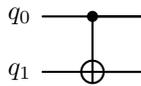
\begin{figure}[ht]
    \centering
    \begin{quantikz}
        \lstick{$q_0$} & \ctrl{1} & \qw \\
        \lstick{$q_1$} & \targ{} & \qw
    \end{quantikz}
    \caption{CNOT gate in the circuit settings. The top qubit $q_0$ is the control, while $q_1$ is the target.}
    \label{fig::graphs-reducts_circ-synth_cnot-gate::cnot-circ}
\end{figure}

CNOT's  graphical counterpart is depicted in Figure \ref{fig::graphs-reducts_circ-synth_cnot-gate::cnot-circ}.

The two qubits on which a CNOT acts are usually referred to as \emph{control} and \emph{target}.
Roughly speaking, the effect of a CNOT is to \emph{flip the target qubit if and only if the control qubit is one.}
The behaviour of a CNOT gate is the same as the xor logical operation, that returns 1 if and only if exactly one of the two arguments is 1.
Another example of a gate that acts in a similar manner is Toffoli, which in turn takes as input three qubits. 
The third one is flipped if and only if both the other two are one. 
Composing CNOT and TOFFOLI allows to obtain circuits that flip the target qubit if and only if the controls hold some specific truth value. 

As we mentioned before, the application of a CNOT ---and, in general, of a multi-qubits gate---must be done carefully.
This because physical implementation of two qubits gates we are provided with are intrinsically faulty and can produce mixed states even if the input state was originally pure. 
Therefore, it is natural to think about reducing the number of two qubits gates---CNOTs---we adopt when we rewrite a unitary in terms of smaller terms. 
The \textbf{CNOT-count problem} arises here: find the implementation of a unitary matrix $U$ with the minimum number of CNOT gates. 

This kind of problem is not important just to minimize the noise inside circuits, but also for its effect on stabilizer circuits. 
These particular circuits are the ones that can be produced just by using gates inside the Clifford group. 
This set is not universal, but it describes perfectly which quantum algorithms can be simulated efficiently with classical hardware \cite{gottesman1998heisenbergrepresentationquantumcomputers}.

A subsequent result about stabilizer circuits comes from \cite{Aaronson_2004} in which authors proved the existence of a 
\emph{normal form}. 
\begin{theorem}[\cite{Aaronson_2004}]
    \label{thm::graphs-reducts_circ-synth_cnot-gate::stab-nf}
    Let $\mathcal{C}$ be a circuit consisting of CNOT, Hadamard and phase gates---a stabilizer. 
    There exists an equivalent circuit with the following normal form: 
    \begin{equation*}
        H-C-P-C-P-C-H-P-C-P-C
    \end{equation*}    
    where an $H, C, $ and $P$ indicates a layer of Hadamard, CNOT, and Phase gates, respectively. 
\end{theorem}
Layers of Hadamard and Phase gates have length exactly one---being single qubit gates, they can all be parallelized.
On the other hand, the whole length of the circuit depends on the length of the CNOT layers. 
Therefore, it is crucial to produce algorithms that can minimize the number of such gates. 

Circuits composed only of CNOT gates are usually called either \emph{CNOT circuits} or \emph{Linear reversible circuits}.
Since the effect of a CNOT gate is intrinsically classical, CNOT circuits can be represented using classical tools.  
In particular, any linear reversible circuit on $n$ qubit can be completely described by an $n \times n$ boolean invertible\footnote{Notice that if the matrix is not invertible, than the circuit cannot be decomposed in terms of CNOT gates.} matrix $M$. 
As the reader may notice, we converted a problem on $n$ qubit---hence, with a representation that has dimension $2^n \times 2^n$---to a problem on a $n \times n$ matrix. 
Because of this shift in dimension, the definition of CNOT we gave in Eq. \ref{eq::graphs-reducts_circ-synth::CNOT} is not coherent anymore. 
\begin{definition}
    Let $M$ be an $n \times n$ matrix representing a linear reversible circuit. The effect of applying a CNOT with $i$-th row as a control and 
    the $j$-th row as a target is to mupliply $M$ by a matrix $M'$ defined as follows.
    $M'$ is the identity matrix with an additional off-diagonal 1 element in position $i, j$.
\end{definition}

This particular definition of the effect of a CNOT gate, allows us to treat it as an elementary row operation. 
\begin{lemma}
    The action of a CNOT gate, whose control is on the $i$-th row and target is in the $j$-th row, is an elementary row operation on a $n \times n$ matrix, which adds the $i$-th row to the $j$-th row.
\end{lemma}
The sum must be interpreted in modulo 2. 
Before formally stating the CNOT minimization problem, we introduce the following theorem: 

\begin{theorem}[\cite{artin2011algebra}]
    Any invertible matrix can be transformed into an identity matrix using elementary row and/or column operations. Thus, any invertible matrix can be decomposed as a product of elementary matrices.    
\end{theorem}

Therefore, any invertible matrix $M$---linear reversible circuit---can be decomposed into the identity matrix using only CNOT gates. 

\begin{definition}
    Let $M$ be an invertible binary matrix. The CNOT minimization problem is the problem of finding the shortest sequence of $k$ matrices $M_1, M_2, \dots, M_k$ such that  $ M M_1 M_2 \dots M_k = \mathbb{I} $ where each $M_i$ is the matrix of a CNOT. 
\end{definition}

Dually, it can also be defined as the problem of turning the identity matrix into $M$ applying only CNOT gates.

Before moving to the complexity topics related to CNOT minimization, we would like to stress one peculiar aspect about CNOT metric. 
In our settings, a CNOT is thought to be applied to rows $i, j$ and the result overwrites one of the two rows. 
With this definition, we are calculating the \emph{s-XOR} metric. 
In the literature, however, \emph{g-XOR} exists as well. In this case, a CNOT takes into account three indices $k, i, j$ and the result of the XOR between $i, j$ is stored in row $k$.

As previously mentioned, we focus on s-xor metric.

To conclude this section, we want to clarify that as stated in \cite{bu2024minimumsynthesiscostcnot}:
\emph{The computational complexity of the CNOT-optimal circuit synthesis (with no constraints on the topology) is yet to be established}.

\subsection{Patel-Markov-Hayes in a Nutshell}

The Patel-Markov-Hayes (PMH) algorithm, introduced in \cite{PatelMarkovHayes}, derives its name from its three authors. While initially not widely adopted, the increasing interest in quantum computing has led to a renewed focus on circuit optimization techniques. The significance of CNOT gate minimization, coupled with the normal form introduced for stabilizer circuits (Theorem \ref{thm::graphs-reducts_circ-synth_cnot-gate::stab-nf}), has established PMH as a state-of-the-art\footnote{Not in terms of gate count: it is the only heuristic-based algorithm whose worst-case performance is proven to match the asymptotic optimum of $\mathcal{O}(n^2/\log n)$.} method for CNOT minimization \cite{PMHOptimal}.

Let $M \in \{0,1\}^{n \times n}$ be an invertible Boolean matrix over the field $\mathbb{F}_{2}$. The PMH algorithm operates by processing the matrix in \emph{stripes} of width $k$, where $k$ is an integer in $\{1, 2, \dots, n\}$. Conceptually, the algorithm partitions $M$ into $\lceil n/k \rceil$ vertical stripes and iteratively processes them. The goal is to transform the first stripe, corresponding to columns $0$ through $k-1$, into an upper triangular form by ensuring:

\begin{itemize}
    \item $M_{i,i} = 1$ for all $i \in \{0, 1, \dots, k-1\}$ (diagonal elements set to 1).
    \item $M_{i, j} = 0$ for all $i \in \{0, 1, \dots, k-1\}, j > i$ (zeroing out elements below the diagonal).
\end{itemize}

This transformation is achieved in two phases. First, rows with identical entries in the first $k$ columns are combined via XOR operations to reduce redundancy. Second, a Gaussian elimination procedure is applied, ensuring that all elements below the diagonal are set to zero while maintaining ones on the diagonal. Once all stripes have been processed, the resulting matrix $M'$ is in an upper triangular form.

The algorithm then transposes $M'$ and reapplies the same procedure. To preserve the triangular structure obtained before transposition, it is crucial that row operations are performed such that the row with the smaller index serves as the control, preventing the introduction of new ones in previously zeroed regions of the matrix.

\section{Method}
\label{sec::methods}

In the previous section we introduced the CNOT-Minimization task. 
Given an invertible boolean matrix $M \in \{0, 1\}^{n \times n}$, we want to turn it into the identity matrix using only elemental row operations---CNOTs.

Both exact and heuristic-based methods struggle to deal with this problem. 
The former class fails to scale when the size of the problem grows \cite{piazza2023asp}.
On the other hand, the latter type produces a gate count that is far from the optimum \cite{PatelMarkovHayes}.

In this paper we propose a Deep Reinforcement-Learning (RL) method to address the CNOT-minimzation problem.
Deep RL approaches are becoming more and more prominent when dealing with circuit synthesis \cite{weiden2024highprecisionfaulttolerantquantum,vanderlinde2023qgymgymtrainingbenchmarking} because of the lack of natural methods to obtain ground truths. This particular fact does not allow the application of supervised learning techniques. 

Our approach to CNOT minimization employs a PPO2~\cite{schulman2017proximal} agent implemented using the \textit{stable-baselines 3} framework~\cite{raffin2021stable}. The agent is trained to iteratively apply CNOT operations to reduce a given matrix to the identity. The agent operates within a discrete action space of $\mathcal{O}(n^2)$ possible moves, where each move is defined as a tuple $(i, j)$, corresponding to the application of a CNOT gate that updates row $i$ by performing an XOR operation with row $j$.

Since an RL agent is trained on a specific state space, it is generally not possible to train a single model on matrices of different sizes. Consequently, a separate model, along with a new training process, is required for each $n$, significantly limiting the approach usability and scalability. To address this limitation, we apply matrix manipulation techniques that enable training a model on a fixed size while allowing the agent to be applied to matrices of different sizes.

\subsection{Curriculum Learning Strategy}

Due to the structural properties of the CNOT minimization problem, we adopt a curriculum learning approach to progressively increase the complexity of the training instances. The training process consists of the following stages:

\begin{enumerate}
    \item \textbf{Permutation Matrices:} The agent first learns to optimize CNOT sequences for permutation matrices, which form a fundamental subset of invertible binary matrices.
    
    \item \textbf{Triangular Matrices:} Once the agent demonstrates proficiency in handling permutation matrices, it advances to upper and lower triangular matrices. These matrices introduce additional structure while remaining relatively tractable.
    
    \item \textbf{Generic Matrices:} In the final stage, the agent is trained on arbitrary invertible binary matrices, ensuring its capability to generalize to complex instances.
\end{enumerate}

As the reader may note, matrices are in some sense \emph{sorted} in terms of complexity. 

Injecting our knowledge about the problem, we assume permutation matrices to be the easier to turn into the identity. 
Subsequently, we place the set of triangular (either upper or lower) matrices. Having one entire portion of the matrix set to 0 is clearly an easier setting than a completely generic matrix.

That being the case, the model is exposed to randomly generated matrices only after the two above mentioned phases are completed. 

\subsection{Gaussian Striping}

As we mentioned before, increasing the size of the matrices handled by the agent require more and more training time. 
Therefore, we propose a preprocessing step that, given a model $\mathcal{A}$ that solves the CNOT minimization problem for dimension $m$, allows to solve instances with $n > m$.
We refer to this process as \emph{Gaussian striping}, since, informally, it edits a stripe of the input matrix adopting a gaussian-like process. 

Let $M \in \{0,1\}^{n \times n}$ be a boolean invertible matrix with $n > m$.
The modification consists of applying a slightly altered PMH-like approach to a stripe of width $k = n - m$, effectively reducing the first $k$ columns into an upper triangular form. The same procedure is then applied after transposing the matrix, resulting in a block-diagonal structure:

\begin{equation*}
    M'' = \begin{bmatrix}
        I_{k} & 0 \\
        0 & M'
    \end{bmatrix},
\end{equation*}

where $I_k$ is the $k \times k$ identity matrix and $M'$ is an $(n-k) \times (n-k)$ invertible Boolean matrix. The transformed matrix $M'$ has its first $k$ columns reduced to an identity block, allowing us to isolate the remaining $M'$ submatrix for further optimization.

The modified PMH procedure is shown in Algorithm \ref{algo::shrinking_algo}.

\begin{algorithm}
\caption{Gaussian Striping Algorithm}
\label{algo::shrinking_algo}
\begin{algorithmic}[1]
    \Function{Reduce}{$M$}
    \State Apply PMH to the first $k = n-m$ columns of $M$.
    \State Transpose $M$.
    \State Apply PMH to the first $k = n-m$ columns of the transposed $M$.
    \EndFunction
\end{algorithmic}
\end{algorithm}

The total CNOT count required to transform $M$ into the identity matrix is the sum of:
\begin{itemize}
    \item The CNOT count required by the preprocessing to produce $M''$, and
    \item The CNOT count required by $\mathcal{A}$ to reduce $M'$ to the identity.
\end{itemize}

Notice that this approach has no scalability issues since PMH tweaked version we apply has complexity $\mathcal{O}(n^{3})$ on a $M \in \{0,1\}^{n \times n}$ matrix.

\subsection{Embedding Small Matrices}

Let $\mathcal{A}$ be an agent capable of solving the CNOT minimization on matrices $M \in \{0, 1\}^{m \times m}$.

For matrices $M' \in n \times n$, with $n < m$ we employ an embedding strategy that maps the problem into an $m \times m$ space while preserving its structure. Given an invertible Boolean matrix $M' \in \{0,1\}^{n \times n}$ with $n < m$, we construct an augmented matrix $M''$:

\begin{equation*}
    M'' = \begin{bmatrix}
        I_{k} & 0 \\
        0 & M'
    \end{bmatrix},
\end{equation*}
where $k = m - n$. This embedding ensures that $M''$ is an $m \times m$ invertible matrix while preserving the original problem structure.

The resulting matrix $M''$ is then given as input to the agent $\mathcal{A}$, which optimizes the CNOT sequence. This approach allows us to unify the treatment of matrices with dimensions up to $m \times m$ while ensuring that smaller instances remain compatible with our RL-based optimization framework.

\section{Experimental Evaluation}
\label{sec::exp_eval}
To evaluate the performance of our proposal, we designed a series of experiments with varying levels of difficulty and different problem sizes. As the problem size increases, so does the complexity of training an RL agent, both in terms of computational cost and the number of required epochs. For this reason, we conducted training on size $m=8$. 
This choice was made for two reasons. First, it enables fast training and evaluation. Second, it has been shown in \cite{piazza2023asp,Meuli2018SATbasedT} that problems of this size begin to become intractable for exact methods and require heuristic approaches.

We compare our method with the original PMH algorithm for two main reasons: (i) it is one of the most widely used—implemented as a native function in \textsc{Qiskit}, (ii) it has provable guarantees about its worst-case performance \cite{Goubault_de_Brugi_re_2022}.

\subsection{Agent Training}

This part of the paper is devoted to the description of the value adopted for the agent's training phase. 
This step took around one and a half day to complete. 

\subsubsection{Training Schedule and Episode Allocation}

As we mentioned before, during the training phase we adopted a curricular approach exposing the model to matrices of growing complexity. 
A curriculum is composed of a certain number of episodes. During each episode, the model is provided with exactly one matrix. 

We now describe the curriculum ranges, in terms of number of episodes and matrix complexity. 
\begin{itemize}
    \item The initial training phase focuses on \emph{permutation matrices} and \emph{triangular matrices}. A total of 3000 episodes are allocated to this phase, evenly distributed among the two classes, with 1000 episodes dedicated to each.
    
    \item From episode 3000 to episode 6000, training continues with matrices randomly sampled from permutation and triangular ones, reinforcing the agent's ability to generalize over these structures.

    \item Between episodes 6000 and 10000, the model transitions to handling \emph{randomly generated matrices}, specifically those created using $n/2$ CNOT operations.

    \item The training phase from episode 10000 to 20000 is dedicated to matrices generated with $n$ CNOT operations, increasing the problem complexity.

    \item More complex cases arise in subsequent stages: matrices generated using $n \log n$ CNOT operations are introduced between episodes 20000 and 50000. Matrices requiring $n^2$ operations are used for the final 50000 episodes.
\end{itemize}

This structured progression ensures that the RL agent is exposed to increasingly complex instances in a controlled manner, facilitating a gradual adaptation from simpler structures to highly populated matrices. By proportionally increasing the number of episodes allocated to more complex cases, we enhance the model's ability to optimize CNOT sequences effectively across different problem instances.

\subsubsection{Reward Function Design}

Our reward function is designed to minimize the introduction of external problem-specific heuristics, ensuring that the model learns an optimized strategy purely through interaction. 

To achieve this, we structure the reward as follows:

\begin{itemize}
    \item A reward of $+0.7$ is assigned whenever the model successfully \emph{solves} a matrix, meaning that it transforms it into the identity matrix.
    \item If the matrix is not fully solved, we compare the states before and after the applied move. Let $M_1$ and $M_2$ denote the matrices before and after the action, respectively.
\end{itemize}

At this point, we distinguish between two cases. 
If the Hamming distance between $M_1$ and the identity matrix differs from that of $M_2$ and the identity, the reward function considers the number of diagonal and off-diagonal ones in both matrices. The assigned reward is $0.2 \times d/n - 0.1 \times d/n^{2}$, where:
\begin{itemize}
    \item $d$ represents the change in the number of ones on the diagonal between $M_1$ and $M_2$.
    \item $\overline{d}$ represents the same difference but for off-diagonal ones.
\end{itemize}

This formulation ensures that the model is penalized for introducing extra off-diagonal ones or removing diagonal ones. Conversely, it is rewarded for eliminating off-diagonal ones and correctly placing ones on the diagonal.

On the other hand, if the applied move does not alter the Hamming distance from the identity matrix, the action is considered unproductive, and a small penalty of $-0.001 / n^2$ is applied.
By structuring the reward function in this way, we encourage the model to prioritize transformations that reduce circuit complexity while avoiding redundant operations.

\subsection{Setting}

To evaluate our approach, we considered sizes ranging from $n=3$ to $n=15$ (corresponding to $3$ and $15$ qubits, respectively). For each size, we generated a set of $100$ matrices by applying a setting-dependent number of CNOT gates. The following settings, increasing in complexity, were used:
\begin{itemize}
    \item \textbf{Rare:} Generated by applying $n/2$ CNOT.
    \item \textbf{Medium:} Generated by applying $n \log n$ CNOT.
    \item \textbf{Overcooked:} Generated by applying $n^2$ CNOT.
\end{itemize}

For each setting, we compare the PMH algorithm with our proposed RL agent, which was trained on matrices of size $8\times 8$. For sizes $n\leq 8$, we applied the agent after matrix embedding the initial instance in an $8\times 8$ matrix. Conversely, for $n\geq 9$, we employed the dimensionality reduction technique, as described in Section \ref{sec::methods}, first reducing the matrix size and then applying the RL agent. In these cases, we report the total number of CNOT gates required, including both the reduction step and the agent’s output.

Finally, to account for the inherent stochasticity of RL-based approaches, we evaluated the agent over $100$ runs, retaining only the best results.

\begin{table*}[htbp]
\begin{center}
\begin{tabular}{ccccccc}
\toprule
 Setting & \multicolumn{2}{c}{\textit{Rare}} & \multicolumn{2}{c}{\textit{Medium}} & \multicolumn{2}{c}{\textit{Overcooked}} \\
Size &  RL &  PMH &  RL &  PMH &  RL &  PMH \\
\midrule
$3^-$ & 2.13 ± 0.34 & 1.61 ± 0.95 & 3.40 ± 1.32 & 3.39 ± 1.41 & 3.60 ± 1.24 & 3.52 ± 1.51 \\
$4^-$ & 3.01 ± 1.15 & 2.84 ± 0.97 & 6.92 ± 8.42 & 5.76 ± 2.07 & 7.19 ± 1.74 & 6.98 ± 1.90 \\
$5^-$ & 2.95 ± 1.01 & 2.84 ± 0.86 & 10.37 ± 2.43 & 10.26 ± 2.35 & 11.84 ± 7.80 & 11.07 ± 2.37 \\
$6^-$ & 4.00 ± 1.18 & 3.98 ± 1.26 & 14.16 ± 5.67 & 14.52 ± 3.38 & 16.20 ± 7.21 & 16.40 ± 2.78 \\
$7^-$ & 4.09 ± 1.08 & 4.03 ± 1.20 & 18.87 ± 9.59 & 19.04 ± 4.35 & 22.61 ± 9.01 & 22.62 ± 3.64 \\
$8$ & 5.25 ± 1.37 & 5.29 ± 1.26 & 22.82 ± 3.52 & 26.27 ± 5.34 & 28.02 ± 6.41 & 30.58 ± 4.07 \\
$9^*$ & 4.29 ± 1.02 & 4.21 ± 1.00 & 32.00 ± 5.26 & 35.55 ± 7.49 & 36.19 ± 5.28 & 40.94 ± 5.36 \\
$10^*$ & 5.63 ± 1.20 & 5.55 ± 1.28 & 40.05 ± 5.87 & 46.22 ± 7.63 & 45.58 ± 7.34 & 52.10 ± 7.04 \\
$11^*$ & 5.64 ± 1.37 & 5.50 ± 1.21 & 48.34 ± 6.87 & 57.18 ± 9.66 & 55.29 ± 8.47 & 66.98 ± 7.57 \\
$12^*$ & 6.86 ± 1.68 & 6.47 ± 1.62 & 58.88 ± 7.21 & 70.42 ± 9.48 & 63.99 ± 5.02 & 82.77 ± 7.80 \\
$13^*$ & 6.98 ± 1.39 & 6.45 ± 1.10 & 66.88 ± 10.15 & 85.82 ± 13.37 & 77.35 ± 6.01 & 103.83 ± 8.26 \\
$14^*$ & 8.17 ± 1.89 & 7.62 ± 1.47 & 79.00 ± 11.86 & 101.41 ± 15.29 & 90.95 ± 7.84 & 123.89 ± 9.61 \\
$15^*$ & 8.43 ± 1.49 & 7.93 ± 1.38 & 86.83 ± 11.50 & 118.46 ± 17.07 & 106.70 ± 8.70 & 148.57 ± 10.04 \\
\bottomrule
\end{tabular}
\end{center}
\caption{Average CNOT Count for the considered sizes and settings. For sizes lower than $8$, denoted with $^-$, the submatrix embedding is used before applying the RL method. For sizes higher than $8$, denoted with $^*$, is applied the Gaussian Stripe algorithm (Algorithm \ref{algo::shrinking_algo}). Values are presented as mean $\pm$ standard deviation.}
\label{tab:cnot_avg}
\end{table*}

\subsection{Evaluation}

\begin{figure}
    \centering
    \includegraphics[width=\linewidth]{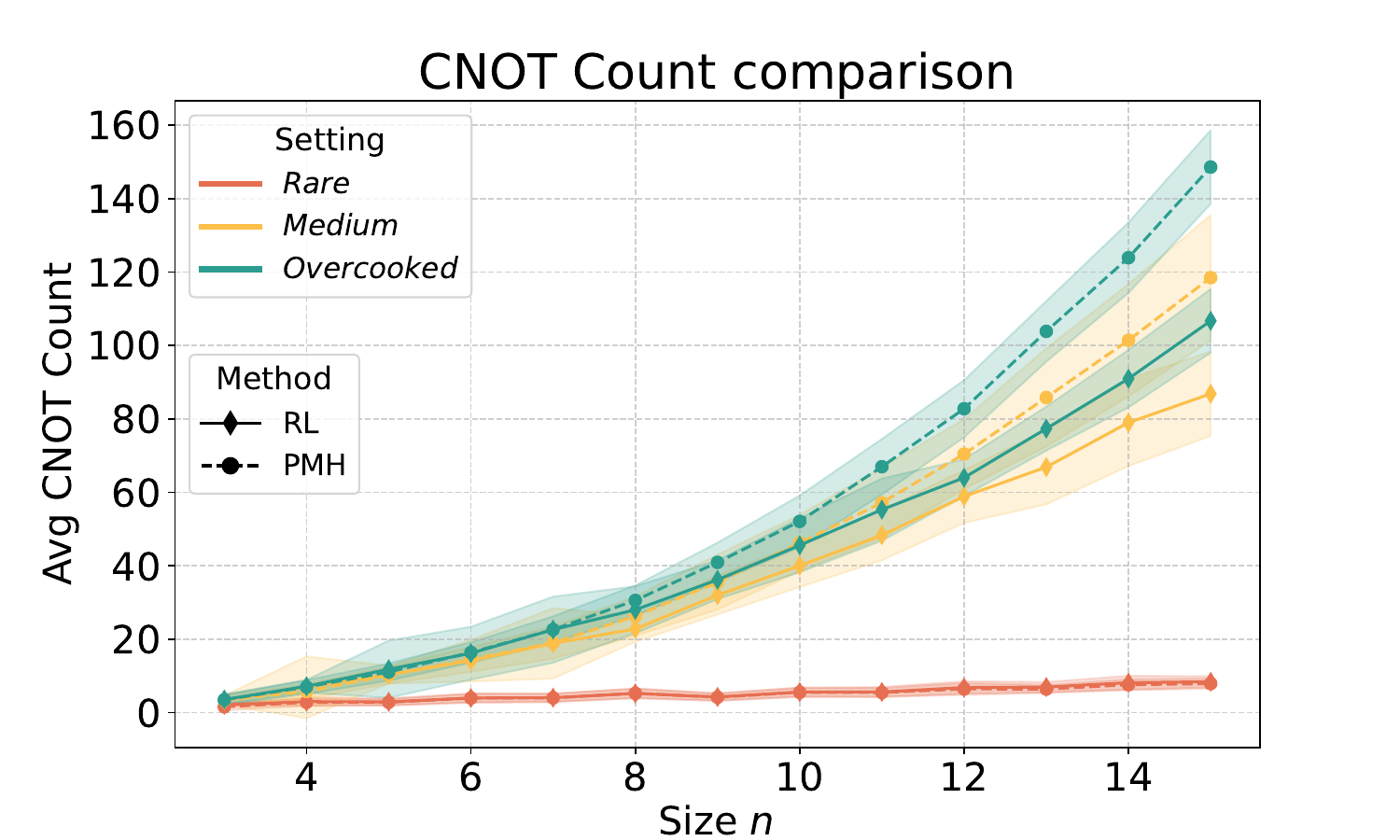}
    \caption{Comparison between the number of CNOT generated by PMH and RL approaches as $n$ grows, for the three different settings considered.}
    \label{fig::line-plot}
\end{figure}

For each setting and matrix size, Table \ref{tab:cnot_avg} reports the average number of CNOT gates required, along with the corresponding standard deviation. A graphical summary of the results is provided in Figure \ref{fig::line-plot}.

An initial observation is that for $n\leq 7$, the average number of CNOT gates required by both approaches remains comparable across all settings. However, as noted in \cite{piazza2023asp,Meuli2018SATbasedT}, for these problem sizes, the task can be exactly solved via logic programming, SAT solvers, etc. Consequently, heuristic-based methods are less relevant in this regime.

For $n=8$, significant differences begin to emerge, particularly in the \textit{Medium} and \textit{Overcooked} settings, where the RL agent achieves an average reduction of approximately $3$ and $2$ CNOT gates, respectively, compared to PMH.

For $n\geq 9$, our proposed method consistently outperforms PMH across all three settings. Notably, this advantage is independent of the specific strategy used to generate the test instances, highlighting the robustness of our approach.

\begin{figure*}[h]
\centering
  \includegraphics[width=0.94773\linewidth]{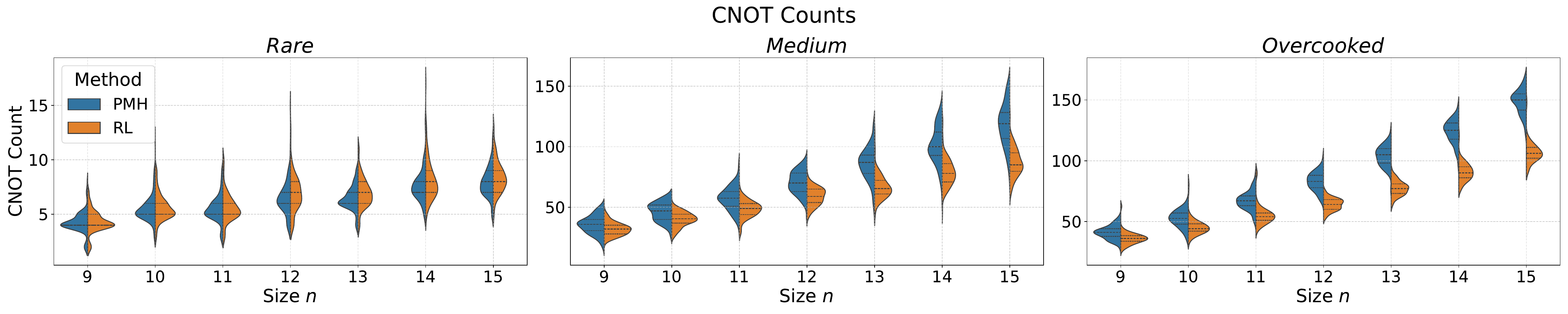}
  \caption{Comparison between the distribution of CNOTs used by PMH and RL for $n\geq 9$, for the settings considered.}
  \label{fig::violin-H}
\end{figure*}

To further analyze the performance for $n \in \{9, \dots, 15 \}$, Figure \ref{fig::violin-H} presents violin plots depicting the distribution of CNOT counts for each setting. In the \textit{Rare} setting, PMH exhibits a few outliers contributing to a higher average CNOT count. However, the overall distributions indicate that our method consistently achieves lower CNOT counts, with a performance gap that widens as $n$ increases.

In the \textit{Medium} and \textit{Overcooked} settings, the difference between the distributions becomes even more pronounced. For higher values of $n$, our approach consistently discovers circuits with a reduced number of CNOT gates. This trend is particularly evident starting from $n=11$ in the \textit{Overcooked} setting, where even the worst-performing solutions generated by our method still require fewer CNOT gates than the best solutions obtained via PMH.

\subsection{Gaussian Striping applied to PMH}
As part of our ablation study, we briefly compare the impact of our proposed method for $n \geq 9$, i.e., in cases where the gaussian striping is applied, with the PMH algorithm on the reduced matrices. 
This comparison aims to determine whether the advantage in CNOT counts arises solely from the matrix manipulation technique or from the RL agent.

We report the difference between the average number of CNOTs used by our method and by the PMH$^*$ algorithm (i.e., PMH applied to the reduced matrices obtained via gaussian striping) in Table \ref{tab:cnot_avg_pmh*}. In the \textit{Rare} setting, the difference between the two approaches is close to zero, while for \textit{Medium} and \textit{Overcooked}, we observe a reduction of approximately $1 \sim 3$ gates. This result is consistent with those presented in Table \ref{tab:cnot_avg_pmh*} for $n=8$, as the advantage of our RL agent depends only on the transformations applied to matrices of the initial size.

\begin{table}[htbp]
\begin{center}
\begin{tabular}{cccc}
\toprule
Size & \textit{Rare} & \textit{Medium} & \textit{Overcooked} \\
\midrule
9 & -0.05 & 3.29 & 2.78 \\
10 & 0.01 & 2.57 & 2.00 \\
11 & 0.02 & 3.08 & 1.41 \\
12 & -0.17 & 2.29 & 3.19 \\
13 & -0.15 & 1.66 & 2.99 \\
14 & -0.21 & 1.24 & 3.01 \\
15 & -0.30 & 3.08 & 2.14 \\
\bottomrule
\end{tabular}
\end{center}
\caption{Difference in number of CNOTs used between our method and PMH$^*$, corresponding to first reduce the matrix to a size $8\times 8$, and then using the standard PMH algorithm.}
\label{tab:cnot_avg_pmh*}
\end{table}

\section{Conclusions and Future Work}
\label{sec::concl}

The gate-based model of quantum computation requires quantum algorithms to be expressed in terms of fundamental operations known as quantum gates. Among these, the CNOT gate plays a crucial role, acting on two qubits and serving as the quantum analogue of the classical XOR operation.

CNOT gates are particularly significant in stabilizer circuits, a fundamental component of Quantum Error Correction. However, due to their multi-qubit nature, they introduce noise into the quantum system, necessitating their minimization. This leads to the \emph{CNOT-minimization problem}, which seeks to transform a given CNOT circuit  into an equivalent one with the minimum number of such operations.

In this work, we proposed a Reinforcement Learning-based approach to address this problem. We trained an RL agent capable of solving instances on 8 qubits---matrices of size $8 \times 8$. 
For circuits with less than 8 qubits, we embedded their representation in a higher dimension matrix amenable to be solved by the RL agent.
On the other hand, for circuits exceeding this size, we introduced a preprocessing step, called Gaussian Striping, that partially resolves the problem by reducing the input instance to an 8-qubit subproblem.

To evaluate our approach, we conducted extensive testing on circuits up to 15 qubits, comparing our results against the Patel-Markov-Hayes (PMH) algorithm, the state-of-the-art method for CNOT minimization. Our findings demonstrate that our RL-based technique outperforms PMH when the number of qubits exceeds 8, highlighting its effectiveness for larger problem instances.

We observed that, when comparing RL with PMH combined with Gaussian Striping, the reduction in CNOT count mainly depends on the agent's performance on matrices of the size used for training. Hence, a promising direction is to optimize the RL pipeline to enable training on larger initial sizes.

Alternatively, another research avenue involves improving the preprocessing step to enhance its efficiency, minimize the number of additional CNOT gates introduced during problem reduction, and potentially extend its applicability to different synthesis approaches. This is supported by the improved performance of PMH after the application of Gaussian Striping.

By advancing in these directions, we aim to enhance the applicability of RL methods for circuit synthesis, contributing to more efficient quantum computing architectures.

\bibliographystyle{IEEEtran}
\bibliography{IEEEexample.bib}

\end{document}